# Machine Learning Approaches on Crop Pattern Recognition

*A Comparative Analysis*

Kazi Hasibul Kabir, Md. Zahiruddin Aqib, Sharmin Sultana
Department of Computer Science and Engineering
East West University, Dhaka, Bangladesh
Email: (kazihasibulkabir, aqibzahiruddin, ssnila1994)
@gmail.com

Shamim Akhter
Department of Computer Science and Engineering
East West University, Dhaka, Bangladesh
Email: dakhter125@gmail.com

*Abstract*—Agriculture activities monitoring is important to ensure food security. Remote sensing plays a significant role for large scale continuous monitoring of cultivation activities. Time series remote sensing data were used for the generation of the cropping pattern. Classification algorithms are used to classify crop patterns and mapped agriculture land used. Some conventional classification methods including support vector machine (SVM) and decision trees were applied for crop pattern recognition. However, in this paper, we are proposing Deep Neural Network (DNN) based classification to improve the performance of crop pattern recognition and make a comparative analysis with two (2) other machine learning approaches including Naïve Bayes and Random Forest.

*Keywords-Crop Pattern;Machine Learning; Remote Sensing; Deep Neural Network; Naïve Bayes; Random Forest*

I. INTRODUCTION

Agricultural activities monitoring including crop pattern generation is inevitable for estimating the agricultural economy of a country. Crop pattern generation requires proper estimation of crop classification and the classification tools needs proper crop related information. On field survey or other traditional data collection methods demand high cost in both time and economy and includes a number of manpowers. Remote Sensing (RS) data has been proven as a more effective way to monitor crop activities as well as provide useful crop related information.

Many developed and developing countries GDP depends on agriculture. Thailand is a country at the center of the Indochinese in Southeast Asia. Agriculture is one of the major sectors contributing to the GDP of Thailand. In 2017, the agricultural contribution to the GDP of this country is 13% which is (million $) 51949 [1]. For such countries, monitoring agricultural activities is very important. Besides the changing pattern of the crops due to environmental degradation like heavy rainfall, flood, drought etc. leads to find a better crop monitoring system. However, differentiation of the pattern of the crops is a hard task because of the variation of irrigation time, temperature, weather, moisture and quality of the soil. So, crop classification craves a reliable system which can recognize or classify a particular crop with variations.

Researchers applied different algorithms to classify crop pattern, such as, K-means algorithms, Support Vector Machine, Decision Trees etc [3]. We have investigated different machine learning approaches including Artificial Neural Network (ANN), Deep Neural Network (DNN), Naive Bayes (NB) and Random Forest (RF) classifiers to trace the best approach on accuracy domain for crop pattern recognition. ANN is trained to feed the patterns into the network. Later, it is able to recognize once the assignation is done. NB classifiers are the probabilistic classifiers. This algorithm is based on the Bayes theorem where a dynamic self-reliance assumption between the features is made. DNN is different from the single hidden layer ANN. DNN supports many hidden layers and each layer consists of some nodes through which the data passes in a multistep process of pattern recognition. So, the empirical term of DNN means more than one hidden layer and each layer of nodes trains on a distinct set of features based on the previous layer's output [2]. RF is a supervised classification algorithm. It creates a forest by creating some random trees. However, creating the forest is a complex task and is not same as constructing the decision tree with information gain or gain index approach. The above mentioned machine learning approaches are implemented and tested, and then a comparative analysis done to find the optimal one for the crop pattern recognition.

II. RELATED WORKS

Researchers are practicing different crop classification methods to improve the results and to provide a better crop monitoring system. Paper [4] is the most recent work related to crop classification which is done by using RF and Object-based classification (OBIA) scheme for produce land use maps of a smallholder agricultural zone. They use optimization techniques to reduce the number of input variables in the data processing to reduce the processing time by maintaining a smart accuracy level (91.7%). In paper [5], the crop classification is done using RF over time series Landsat 7 ETM+ data. They have used a time series of medium spatial resolution enhanced vegetation index (EVI) to produce variables for the RF classifier. Susceptibility of training data, number of variables and accuracy are the evaluation area of their study. They achieved 81% accuracy level. In paper [6], the author implemented different supervised machine learning approaches for crop classification from Multi-Temporal RS image including Softmax Regression-92.06% accuracy, Support Vector Machines (SVMs)-92.21%, Neural Network (NN)-92.06%, and Convolution Neural Network (CNN)-92.64%. In paper [7], the classification of crops is being done using Fuzzy C-Means (FCM) segmentation and texture, color features. In this paper the author used ten different JPEG images from the field. After the segmentation, they have applied the color and texture feature to get the accurate color texture feature of each crop. Thereafter they used Euclidean distance algorithm to identify the crops. ANN classified the crops of rice, sugarcane and maize successfully in [15] with around 70%-75% accuracy. Back propagation is executed for training the data, delta rule is applied to adjust weights. With the enough training data, the ANN can easily encountered the data and recognize them. All of the above mentioned papers applied different machine learning approaches; however, none of them have applied DNN in crop pattern recognition. DNN is not yet been expose in crop pattern generation or classification. Thus, we choose and apply this new machine learning algorithm along with RF and NB and do a comparative analysis among their performance.

### III. STUDY AREA AND REFERENCE DATA

#### A. Study Area

The target area of the study is Suphanburi, located in the Central Plain of Thailand between latitude $14^0 2'43.72"$ and $15^0 5'7.72"$N, longitudes $99^0 17'21.56"$ and $100^0 17'21.56"$E. Different vegetables and crops like maize, sugarcane etc are produced in this province. Rice is the main crop, evolved a few times in a year [15].

#### B. Referenced Data

The experimental data is collected from a report [15]. Suphan Buri 2001-2003 of remote sensing Modis data is used in the study. Moderate Resolution Imaging Spectroradiometer (Modis) data is used with three spatial resolutions: 250, 500, and 1000m. About 44 standards of Modis data products are used which include oceanography, biology and atmospheric science. Users who have proper x-band receiving system may capture the regional data. Different libraries are implemented to process NDVI data[22][24][25]. Modis 8-days NDVI and Modis 32-days NDVI data are calculated through the ENVI format of the images. There are several steps for performing 32-day fusion data. Firstly, re-project the images to lat/long geographic projection system. Secondly, only the study area has been taken as a subset from the image. It covers latitudes $13^0 39'1.72"-16^0 38'43.72"$N and longitudes $98^0 50'21.56"$ - $101^0 50'3.56"$E. From each dataset NDVI is calculated. Lastly, eliminate the clouding pixels using LMF tool. The NDVI image data is also validated by Advanced Very High Resolution Radiometer (AVHRR) data received from NOAA. Some noises are shown in signatures of NDVI image. Quality Controls layer were introduced to discard the "tainted" pixels and enhance the sample by taking more data. But the high level of noise still remains in the signature. Local Maximum Fitting algorithm [21][23] is introduced to NDVI data which creates a noise/cloud free data. After that the Artificial Neural Network (ANN) is applied into the data. The following figure 1 show the land area of Suphan Buri where the different crops are identified.

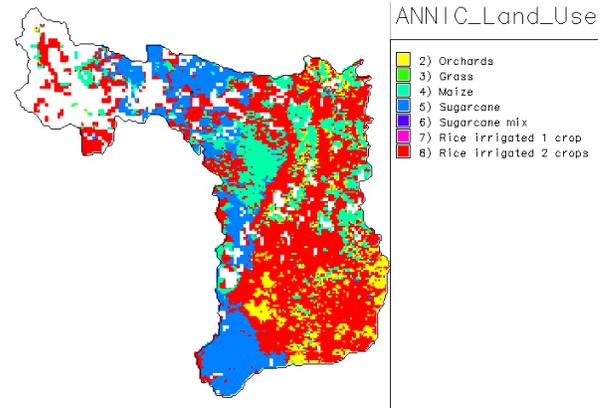

Fig.1. ANN classified land use map.

### IV. IMPLEMENTATION METHODS

In this section we are going to explain the implementation of the machine learning approaches including Naive Bayes(NB), Deep Neural Network (DNN) and Random Forest (RF).

#### A. Naive Bayes Algorithm

Bayesian approach is a supervised learning technique and uses statistical model for classification. This classification is used as a probabilistic learning method like naive Bayes classifier [8]. In a learning problem, naive Bayes classifier requires a number of parameters linearly in the number of variables (features/predictors). It creates a model which assigns class labels to problem instances, represented as vectors of feature value [8]. In this model, class labels are drawn from finite set. NB classifiers assume that given the class variable the value of a particular feature is independent based on the value of any other feature. For example, if rice has some features like white, round etc. NB considers each of these features independently to find out the probability that it is rice. Bayes theorem general equation [9] is,

$$P(C_k|x) = \frac{P(C_k)P(x|C_k)}{P(x)} \quad \ldots (1)$$

$P(C_k | x1,x2,\ldots,xn)$ represents each of K possible outcomes for classes $C_k$. In Eq. (1), the numerator is only significant because the denominator doesn't rely on C [9]. This numerator is equivalent to the joint probability model and after that we can assume each feature $x_i$ is conditionally independent of every other feature $x_j$ for $j \neq i$. So, under this assumption, the conditional distribution over class variable C is,

$$P(C_k|x_1, x_2, \ldots, x_n) = \frac{1}{Z} P(C_k) \prod_{i=1}^{n} P(x_i|C_k) \quad \ldots (2)$$

Where, Z is the evidence. So, if we replace Eq.(2) with y[9],

$$\bar{y} = \frac{1}{z} P(C_k) \prod_{i=1}^{n} P(x_i|C_k) \qquad \text{...(3)}$$

To run the NB model, there are some defined parameters. First of all, for evaluating the model, dataset is being split in training dataset (70%) and test dataset (30%). The optimization was processed by naivebayes function which was found in library caret package available in the R software. In the naivebayes function, factor variable is declared and training dataset is defined.

### B. Deep Neural Network (DNN)

Deep learning or deep neural network is a recent topic of machine learning which models the non-linear representations of data. In previous, it is hard to train artificial neural network because of vanishing gradients and over fitting problems [13]. In deep learning both of these problems has been solved by different activation functions and dropout regularization and a huge amount of data. In deep neural network, multiple layers are being used. Each successive layer uses the output from the previous layer as input. There are many methods to implement deep neural network. Softmax regression and Rectifier Linear Unit (ReLU) are two of them. Softmax regression is a logistic regression which can handle multiple classes. The equation of softmax regression is,

$$\sigma(z)_j = \frac{e^{z_j}}{\sum_{k=1}^{K} e^{z_j}} \qquad \text{...(4)}$$

Where, j =1,2,..... K

The output of softmax regression can be used as a probabilistic distribution over K different possible outcomes [14]. ReLU is an activation function which can be defined as, $f(x)= \max(0,x)$ where, x is the input of the neuron. To evaluate the deep neural network model, keras (create tensor flow environment) library will be needed for R software [14]. In the model three layers are created. First layer is the input layer with 136 neurons, second and third layers are the hidden layers with 68 neurons and 32 neurons respectively, and the final layer is the output layer with 8 neurons. In the model, ReLU activation function is set for input layer and hidden layer. For calculating the prediction output, softmax regression is set to the model. There are 8 class variables so that the output neuron is eight (8). By optimizing 136 input patterns in the ReLU function, 68 new patterns are created for hidden layer neuron. After that, it predicts the crop by calculating the softmax regression in the output layer. This method is repeated 200 times (epoch = 200). After the iterations, the model calculates the average accuracy based on output prediction.

### C. Random Forest (RF) Algorithm

For classification and regression random forest algorithm is ensemble learning method which consists of a group of tree-based classifiers. It generates an infinitude random decision trees which then aggregated to compute a classification. For crop pattern classification, random forest is found more efficient. Because it takes a few user-defined parameters and return accuracy level which are comparable or better than Decision Tree [18], Support Vector Machine [19] and Neural Network [20]. To build a forest classification trees (300 trees), RF algorithm randomly samples the dataset with replacement. It uses a bootstrap method to enhance the ability of a classification tree. Bootstrap sample is constructed by each tree to find out the Out-of-Bag (OOB) error. Observations of the main dataset that are left out from the bootstrap sample are declared as Out-of-Bag observations [11]. OOB error is calculated using the one-third data from the constructed trees and reciprocates to the rate misclassified variables.

To run the model of RF algorithm, there are some parameters need to be defined. The main parameters of this model are ntree and mtry. Prediction performance of model will be higher if the value of mtry is increased [10]. So for the model optimization of the parameters mtry and ntree are needed to be considered. First of all, for evaluating the model, dataset is being split in training dataset (70%) and test dataset (30%). This splitting is repeated ten (10) times to estimate prediction performance. Each time the training and testing data are used to find out the accurate accuracy by averaging all predictions and estimate each variables importance in the classification.

In the model, there are 300 classification trees and mry is 8. To run this model, class variable, training data and testing data is declared in randomForest function. For R software library randomForest has packages like randomForest which first create tree and forest and then train that forest by training dataset (70%). After training the model, it tests the forest by testing dataset and predicts the crop by evaluating the pattern. These three algorithms calculate the accuracy, Kappa, Sensitivity, Specificity (These are probabilistic statistical measure) by their prediction value. The equations of Kappa, Sensitivity, and Specificity's are discussed below.

- Cohen's Kappa or Kappa is a statistical measure for categorical items. The Kappa statistical value is from 0 to 1 [16]. Most statistical software like R software can calculate Kappa. The formula of Kappa is,

$$Kappa = \frac{P_0 - P_e}{1 - P_e} \qquad \text{...(5)}$$

Where, $P_0$ = Actual probability and $P_e$ = Random probability.

- Two other statistical measure performances are sensitivity and specificity [17]. These two are binary classification test. Sensitivity is known as true possibility rate that measures the probability of positives which are correctly identified by the model [17]. On the other hand, specificity is also known as true negative rate that measures the probability of negativity which are correctly identified [17]. The formulas of these two statistical measures are,

$$Sensitivity = \frac{TP}{TP + FN} \qquad \text{...(6)}$$

$$Specificity = \frac{TN}{TN + FP} \qquad \text{...(7)}$$

Where, TP = True Positive, FP = False Positive, TN

= True Negative, FN = False Negative

TABLE 1. KAPPA'S STATISTICAL VALUE

| 0 | Equivalent to chance |
|---|---|
| 0.1 to 0.20 | Slight probability |
| 0.21 to 0.40 | Fair probability |
| 0.41 to 0.60 | Moderate probability |
| 0.61 to 0.80 | Substantial probability |
| 0.81 to 0.99 | Nearly perfect probability |
| 1 | Perfect probability |

TABLE 2. ACCURACY OF MACHINE LEARNING METHODS

| Algorithm Name | Accuracy (%) |
|---|---|
| Naïve-Bayes (NB) | 60.53 |
| Deep Neural Network (DNN) | 84.89 |
| Random Forest (RF) | 94.74 |

TABLE 3: KAPPA CO-EFFICIENT FOR MACHINE LEARNING METHODS

| Algorithm Name | Kappa (%) |
|---|---|
| Naïve-Bayes (NB) | 62.01 |
| Deep Neural Network (DNN) | 82.19 |
| Random Forest (RF) | 92.88 |

## V. RESULTS AND ANALYSIS

### A. Classification Accuracy

We have carried out the implementation of R programming language using RStudio. All the programs were run in 64-bit Intel core i5 2.5 GHz machine. We have used medium spatial resolution data for our experiment. Our data is collected from Remote Sensing Images. At the beginning of the implementation we divide the whole data set into two parts. One is the training data set which holds 70% of the full data and the other part is test data which is the 30% of the full data. Several libraries and packages have been used in RStudio to implement these algorithms. All the algorithms show different result which is basically statistical measurement like the overall accuracy, Kappa, sensitivity and specificity.

From Table 2, we can see that the accuracy of naive Bayes is 60.53% which is the lowest accuracy. In naive Bayes, the accuracy is calculated based on evidence Z, where class/feature variable is dependent on the input variable $X_i$.
From this table, we can also observe that deep neural network (DNN) gives an accuracy of 84.89%. Two hidden layers, an input layer and an output layer is used in DNN. Two hidden layers provide better accuracy than increasing the hidden layer numbers. This scenario varies from problem to problem. In input layer ReLU (Rectifier Linear Unit) function is used to find the probability of desire output. Normally DNN performs best if there is huge dataset. Our dataset is limited and not sufficient for DNN to provide the best accuracy. Besides the dataset was generated long time ago. The updated dataset may perform well. However, our model is ready to port any amount of data sets.

Random Forest (RF) algorithm is also applied on the limited data sets and acquires 94.74% accuracy. In RF model we create 300 classification trees which are called ntree. Eight (8) prediction/class variables through mtry are implemented. Then the RF is trained by training data. Thereafter, the test features are applied and used the rules of each randomly decision tree to predict the outcome and store them. Then the votes are calculated for each predicted target and consider the high voted prediction target as the final prediction.

To measure the crop identification performance, the Kappa (Cohen's Kappa) co-efficient is calculated (in Table 3) based on accurate accuracy and random accuracy. In NB algorithm, the Kappa value is 62.01% which falls into the substantial probability (Table 1). In DNN, the Kappa accuracy is 82.19% which is nearly perfect probability and in RF, the Kappa probability is 92.88% which also falls into nearly perfect probability.

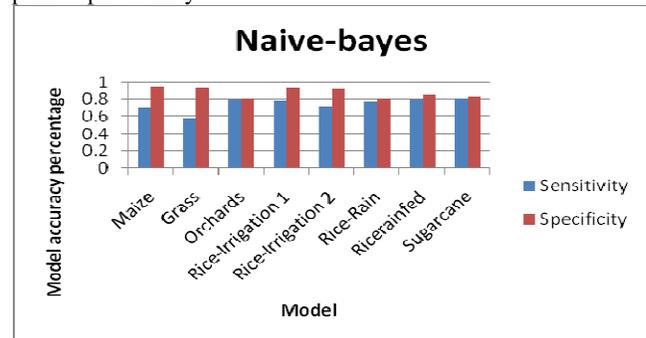

Fig. 2: Sensitivity and Specificity model of Naïve-bayes

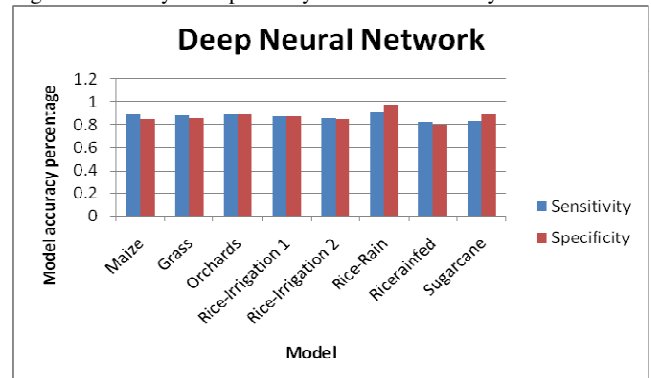

Fig. 3: Sensitivity and Specificity model of Deep Neural Network

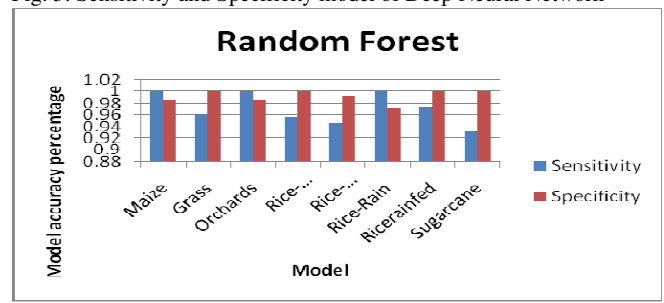

Fig. 4: Sensitivity and Specificity model of Random Forest

The sensitivity probability reflects the correctness of the assumption and the specificity reflects the probability to recognize a correct crop. Fig.2, Fig.3 and Fig.4 shows the corresponding sensitivity and specificity of the crops for Naïve-bayes, Deep Neural Network, and Random Forest machine learning approaches.

*B. Optimization Method*

From the above observation, we can see that the statistical measurement of deep neural network and random forest are very imminent. If we analyze the Kappa accuracy, we can see both the accuracy falls into same range (0.81 to 0.99) which is nearly perfect probability. Although there are some criteria for which the deep learning shows less accuracy than random forest. These are,

- Insufficient observation
- High data dense

Besides random forest works better for some reasons also. The main reason is if there are any missing values, the random forest algorithm calculates an Out-of-Bag (OOB) error and then again uses them for classification. From above discussion, although RF is giving higher accuracy, we prefer deep neural network algorithm for crop classification as it is a newly arrival method and it has a better future scope in the field of classification research.

## VI. CONCLUSION

This study examines that DNN algorithm shows the better result and can be a prominent crop pattern classification method. More robust data is used to get the appropriate classification [15]. Three types of algorithms are implemented - NB, DNN, RF. Among the algorithms, though RF manifests the better performance and highest accuracy (94.74%), DNN performs well (84.89%) in the circumstances to classifying the crops. If we could train the DNN with modern and larger dataset, the accuracy would be gained higher. Besides to train the DNN a huge amount of dataset is needed. The theory behind the algorithms implemented has been correctly handled with proper care. Data has been prepared perfectly and missing values are trained appropriately to handle the errors.

Our main target of this research is to train the machine with different algorithms and find out which algorithms can perform better in a certain situation. With these types of research DNN can be utensil. To locate and identify the rotten crops or which crop is produced well in the certain location can be helpful with our findings. DNN classify the data properly and can distinguish the main crop in the study area. This approach can be applied in the larger area where the data is sufficient enough. However, the process is utilized to gain the highest possible outcome.